%% file: 0_main.tex
\documentclass[a4paper, conference, 10pt]{IEEEtran}

\pdfoutput=1

\IEEEoverridecommandlockouts

\input{01_config}
\input{02_notation}
\input{03_glossary}

\makeatletter
\def\ps@IEEEtitlepagestyle{%
  \def\@oddfoot{\mycopyrightnotice}%
}
\def\mycopyrightnotice{%
\begin{minipage}{\textwidth}
\centering \footnotesize
Copyright~\copyright~2024 IEEE. Personal use of this material is permitted.  Permission from IEEE must be obtained for all other uses, in any current or future media, including reprinting/republishing this material for advertising or promotional purposes, creating new collective works, for resale or redistribution to servers or lists, or reuse of any copyrighted component of this work in other works.
\end{minipage}
}
\makeatother

\begin{document}

\bstctlcite{IEEEexample:BSTcontrol}

\title{
Enhancing Split Computing and Early Exit Applications through Predefined Sparsity
\thanks{
This study was carried out within the PNRR research activities of the consortium iNEST (Interconnected North-Est Innovation Ecosystem) funded by the European Union Next-GenerationEU (Piano Nazionale di Ripresa e Resilienza (PNRR) – Missione 4 Componente 2, Investimento 1.5 – D.D. 1058  23/06/2022, ECS\_00000043), and by the European Union’s Horizon Europe research and innovation programme under the Marie Sklodowska-Curie grant agreement No. 101109243.
This manuscript reflects only the Authors’ views and opinions.
Neither the European Union nor the European Commission can be considered responsible for them.
This work was also partially supported by the US NSF grant 2038960.
}
}

\author{
\IEEEauthorblockN{
Luigi Capogrosso\IEEEauthorrefmark{1},
Enrico Fraccaroli\IEEEauthorrefmark{1}\IEEEauthorrefmark{2},
Giulio Petrozziello\IEEEauthorrefmark{3},
Francesco Setti\IEEEauthorrefmark{1},\\
Samarjit Chakraborty\IEEEauthorrefmark{2},
Franco Fummi\IEEEauthorrefmark{1},
Marco Cristani\IEEEauthorrefmark{1}
}
\IEEEauthorblockA{\IEEEauthorrefmark{1}{\emph{Dept. of Engineering for Innovation Medicine}, University of Verona, Italy, \tt name.surname@univr.it}}
\IEEEauthorblockA{\IEEEauthorrefmark{3}{\emph{Dept. of Computer Science}, University of Verona, Italy, \tt name.surname@studenti.univr.it}}
\IEEEauthorblockA{\IEEEauthorrefmark{2}{\emph{Dept. of Computer Science}, University of North Carolina at Chapel Hill, USA, \tt samarjit@cs.unc.edu}}
}

\maketitle

\begin{abstract}
In the past decade, \glspl{dnn} achieved state-of-the-art performance in a broad range of problems, spanning from object classification and action recognition to smart building and healthcare.
The flexibility that makes \glspl{dnn} such a pervasive technology comes at a price: the computational requirements preclude their deployment on most of the resource-constrained edge devices available today to solve real-time and real-world tasks.
This paper introduces a novel approach to address this challenge by combining the concept of predefined sparsity with \gls{sc} and \gls{ee}.
In particular, \gls{sc} aims at splitting a \gls{dnn} with a part of it deployed on an edge device and the rest on a remote server.
Instead, \gls{ee} allows the system to stop using the remote server and rely solely on the edge device's computation if the answer is already good enough.
Specifically, how to apply such a predefined sparsity to a \gls{sc} and \gls{ee} paradigm has never been studied.
This paper studies this problem and shows how predefined sparsity significantly reduces the computational, storage, and energy burdens during the training and inference phases, regardless of the hardware platform.
This makes it a valuable approach for enhancing the performance of \gls{sc} and \gls{ee} applications.
Experimental results showcase reductions exceeding $4\times{}$ in storage and computational complexity without compromising performance.
The source code is available at \url{https://github.com/intelligolabs/sparsity_sc_ee}.
\end{abstract}


\glsresetall

\input{srcs/1_intro}
\input{srcs/2_related}
\input{srcs/3_method}
\input{srcs/4_experiments}
\input{srcs/5_conclusion}

\bibliographystyle{IEEEtran}
\bibliography{bibi}

\end{document}

%% file: 01_config.tex
\usepackage[T1]{fontenc}
\usepackage[english]{babel}
\usepackage{textcomp}
\usepackage{xspace}
\usepackage{graphicx}
\usepackage{amsfonts}
\usepackage{amsmath}
\usepackage{booktabs}
\usepackage[acronym]{glossaries}
\usepackage[hidelinks, bookmarksopen=true]{hyperref}
\usepackage[capitalize,noabbrev]{cleveref} 
\usepackage{tikz}
\usepackage{pgf}
\usepackage{pgfplots}
\usepackage{subfigure}
\usepackage{multirow}
\usepackage{multicol}
\usepackage{colortbl}
\usepackage{siunitx}

\glsdisablehyper
\AtBeginDocument{}
    
\makeatletter
\newcommand\notsotiny{\@setfontsize\notsotiny\@vipt\@viipt}
\newcommand*{\toroman}[1]{\expandafter\@slowromancap\romannumeral #1@}
\makeatother
\newcommand\mperiod[1][\rlap]{#1{\;.}}
\newcommand\mcomma[1][\rlap]{#1{\;,}}

\newcommand{\tikzxmark}{%
\tikz[scale=0.23] {
    \draw[-, scale=0.6, line width=0.7, line cap=round] (0,0) to [bend left=6] (1,1);
    \draw[-, scale=0.6, line width=0.7, line cap=round] (0.2,0.95) to [bend right=3] (0.8,0.05);
}}
\newcommand{\tikzcmark}{%
\tikz[scale=0.23] {
    \draw[-, scale=0.6, line width=0.7, line cap=round] (0.25,0) to [bend left=10] (1,1);
    \draw[-, scale=0.6, line width=0.8, line cap=round] (0,0.35) to [bend right=1] (0.23,0);
}}

\makeatletter
\DeclareRobustCommand\onedot{\futurelet\@let@token\@onedot}
\def\@onedot{\ifx\@let@token.\else.\null\fi\xspace}
\def\eg{e.g\onedot} 
\def\ie{i.e\onedot}

\def\etal{\emph{et al}\onedot}
\makeatother

\definecolor{red}    {HTML}{b7211f}
\definecolor{orange} {HTML}{FFA500}
\definecolor{blue}   {HTML}{4169E3}
\definecolor{green}  {HTML}{147546}
\definecolor{purple} {HTML}{92268F}

\usetikzlibrary{fit}
\usetikzlibrary{math}
\usetikzlibrary{calc}
\usetikzlibrary{arrows}
\usetikzlibrary{arrows.meta}
\usetikzlibrary{shapes.arrows}
\usetikzlibrary{shapes.symbols}
\usetikzlibrary{shapes.geometric}
\usetikzlibrary{patterns}
\usetikzlibrary{patterns.meta}
\usetikzlibrary{positioning}
\pgfplotsset{compat=newest}
\makeatletter
\tikzset{%
    >=stealth,
    ultra thin/.style= {line width=0.1pt},
    very thin/.style=  {line width=0.2pt},
    thin/.style=       {line width=0.4pt},
    semithick/.style=  {line width=0.6pt},
    thick/.style=      {line width=0.8pt},
    very thick/.style= {line width=1.2pt},
    ultra thick/.style={line width=1.6pt},
    fblue/.style={fill=blue!50},
    fred/.style={fill=red!50},
    forange/.style={fill=orange!50},
    fgreen/.style={fill=green!50},
    fgray/.style={fill=gray!50},
    lblue/.style={draw=blue},
    lred/.style={draw=red},
    lorange/.style={draw=orange},
    lgreen/.style={draw=green},
    lgray/.style={draw=gray},
    AN/.style={anchor=north},
    ANW/.style={anchor=north west},
    ANE/.style={anchor=north east},
    AE/.style={anchor=east},
    AW/.style={anchor=west},
    AS/.style={anchor=south},
    ASW/.style={anchor=south west},
    ASE/.style={anchor=south east},
    AC/.style={anchor=center},
	opaque node/.code 2 args={\tikzset{opacity=#1, text opacity=#2}},
	double color fill/.code 2 args={%
		\pgfdeclareverticalshading[%
		tikz@axis@top,tikz@axis@middle,tikz@axis@bottom%
		]{diagonalfill}{100bp}{%
			color(0bp)=(tikz@axis@bottom);%
			color(50bp)=(tikz@axis@bottom);%
			color(50bp)=(tikz@axis@middle);%
			color(50bp)=(tikz@axis@top);%
			color(100bp)=(tikz@axis@top)%
		}%
		\tikzset{%
			shade,%
			left color=#1,%
			right color=#2,%
			shading=diagonalfill%
		}%
	}%
}
\makeatother

%% file: 02_notation.tex
\newcommand{\layers}{\ensuremath{L}}
\newcommand{\neurons}{\ensuremath{N}}

\newcommand{\indegree}{\ensuremath{d^{in}}}
\newcommand{\outdegree}{\ensuremath{d^{out}}}
\newcommand{\edges}[1]{\ensuremath{|W_{#1}|}}
\newcommand{\density}{\ensuremath{\rho}}

%% file: 03_glossary.tex
\newacronym{sc}{SC}{Split Computing}
\newacronym{ee}{EE}{Early Exit}
\newacronym{mtl}{MTL}{Multi-Task Learning}
\newacronym{stl}{STL}{Single-Task Learning}
\newacronym{dnn}{DNN}{Deep Neural Network}
\newacronym{loc}{LoC}{Local-only Computing}
\newacronym{roc}{RoC}{Remote-only Computing}
\newacronym{losc}{LO-SC}{Local-Only Split Computing}
\newacronym{cnn}{CNN}{Convolutional Neural Network}
\newacronym{rnn}{RNN}{Recurrent Neural Network}
\newacronym{sgd}{SGD}{Stochastic Gradient Descent}
\newacronym{ml}{ML}{Machine Learning}
\newacronym{fc}{FC}{Fully Connected}
\newacronym{svm}{SVM}{Support Vector Machine}
\newacronym{relu}{ReLU}{Rectified Linear Unit}
\newacronym{va}{VA}{Ventricular Arrhythmia}
\newacronym{scd}{SCD}{Sudden Cardiac Death}
\newacronym{milp}{MILP}{Mixed-Integer Linear Problem}
\newacronym{mlp}{MLP}{MultiLayer Perceptron}
\newacronym{ics}{ICS}{Industrial Control System}
\newacronym{it}{IT}{Information Technology}
\newacronym{ot}{OT}{Operational Technology}
\newacronym{llm}{LLM}{Large Language Model}
\newacronym{convnet}{ConvNet}{Convolutional Neural Network}
\newacronym{gpu}{GPU}{Graphical Processing Unit}
\newacronym{tpu}{TPU}{Tensor Processing Unit}
\newacronym{mcu}{MCU}{Micro-Controller Unit}
\newacronym{fpga}{FPGA}{Field Programmable Gate Array}

%% file: srcs/1_intro.tex
\section{Introduction}
\label{cha:intro}

Remarkable improvements in areas like computer vision, speech recognition, and autonomous systems are significantly supported thanks to the rise of \glspl{dnn}~\cite{pouyanfar2018survey,dong2021survey}.
At the same time, over the last decade, \glspl{dnn} have grown significantly in size and complexity; some now have millions or even billions of trainable parameters~\cite{smith2022using}.
However, these powerful models have a significant drawback: they demand substantial computational resources and storage, making deploying them directly on edge devices impractical.

Nowadays, the most commonly used approach is to train huge \glspl{dnn} in server computing settings using high-performance hardware accelerators like \glspl{gpu}~\cite{chan2023deep}, and \glspl{tpu}~\cite{jouppi2017datacenter}.
Then, once trained, the \gls{dnn} model is used for inference, a less computationally expensive activity that, in the case of real-time and real-world problems, is often performed on edge devices, and more in general on embedded systems~\cite{singh2023edge}.

\paragraph*{\textbf{Motivations for this paper}}
As a result, a significant amount of research effort has been directed toward improving embedded technologies.
This focus has enabled the development of real-time solutions for a wide range of real-world complex applications.
In this regard, hardware-specific (\eg{}, edge \glspl{tpu}) and \gls{mcu}-based embedded systems have earned a lot of attention, primarily due to their low power requirements and high performance, and secondarily for their maintainability, adaptability, and reliability~\cite{capogrosso2024machine}.

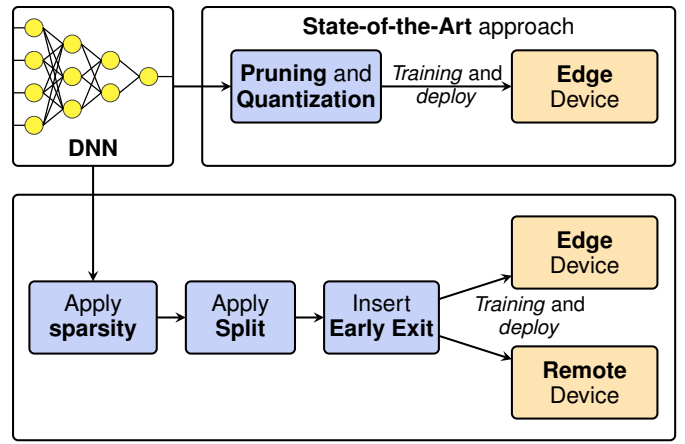
\begin{figure}[!t]
    \centering
    \resizebox{\columnwidth}{!}{\input{figures/front}}
    \caption{Difference between our approach of predefined sparsity applied to \gls{sc} and \gls{ee}, against the state-of-the-art pruning and quantization.}
    \vspace*{-0.5cm}
    \label{fig:cover}
\end{figure}

In particular, the importance of \textit{model reduction} techniques for accelerating \glspl{dnn} is widely acknowledged nowadays, aiming to optimize their performance~\cite{menghani2023efficient}.
These approaches focus on reducing the number of trainable parameters within a \gls{dnn}, resulting in substantial advantages across various aspects: computational resources needed, storage requirements, and energy consumption.
Since efficiency is paramount, improvements in this area are crucial if we aim to deploy \glspl{dnn} on edge devices for solving real-world applications~\cite{fragkou2023model}.

At the same time, current state-of-the-art approaches for efficient edge computing rely on advanced \gls{ml} paradigms, such as \gls{sc} and \gls{ee}~\cite{eshratifar2019jointdnn,jankowski2020joint,sbai2021cut,cunico2022split,capogrosso2023split}.
In particular, \gls{sc}, where a \gls{dnn} is intelligently split with a part of it deployed on an edge device, and the rest on a remote server, and \gls{ee}, where the model is built with multiple ``exits'' across the layers and each exit can produce the model output, represents the state-of-the-art framework for structuring distributed deep learning applications~\cite{matsubara2022split}.
These approaches allow \glspl{dnn} to be leveraged for latency-sensitive applications in scenarios where deploying the entire \gls{dnn} remotely is impractical due to limited local computation bandwidth, and also locally since \gls{dnn} would require higher memory requirements than those available on the edge device.
They combine local and remote computing advantages, leading to lower latency and, more importantly, drastically reducing the required transmission bandwidth.

\paragraph*{\textbf{Innovations in this paper}}
\begin{figure*}[!t]
    \centering
    \input{figures/methodology}
    \caption{Starting from a \gls{dnn} $\mathcal{M}(\cdot{})$, we first apply the \textit{predefined sparsity}, and then we train the network.
    After the training stage, we split the network following the \gls{sc} and \gls{ee} paradigm.
    As a result, the final architecture is not so computationally intensive, doesn't require huge storage spaces, and has less energy consumption, all without compromising the overall performance.}
    \label{fig:methodology}
    \vspace*{-0.5cm}
\end{figure*}
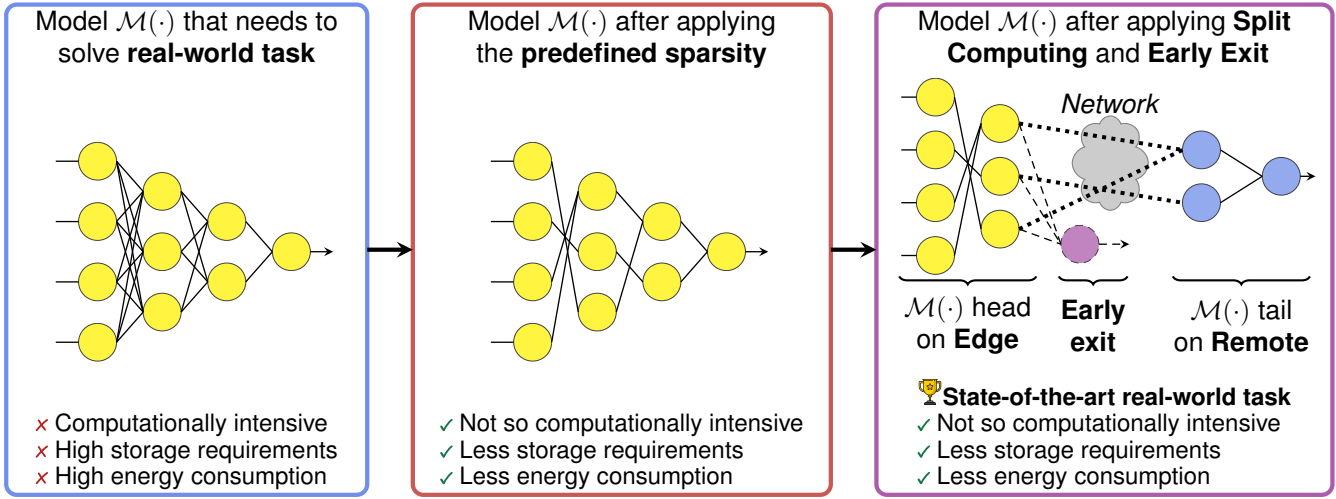
\cref{fig:cover} exemplifies the proposal of this paper, where we explore the application of \textit{predefined sparsity} as a model reduction method within the context of \gls{sc} and \gls{ee}.
We aim to showcase how, in the context of \gls{sc} and \gls{ee}, predefined sparsity can significantly reduce computational demands, storage requirements, and energy consumption compared to state-of-the-art approaches.
Furthermore, regardless of the underlying implementation platform, our approach yields advantages for both the training and inference stages.

In particular, our pipeline is exemplified by the flow presented in \cref{fig:methodology}.
Our strategy for producing sparsity involves defining a preset set of sparse neuron connections before the training process, \ie{}, we eliminate a specified set of connections inside the neural network, and we keep this configuration constant throughout both the training and inference phases.
Specifically, we apply this strategy within an \gls{sc} and \gls{ee} scenario, illustrating how predefined sparsity can enhance performance in applications operating within this context.

As a killer application where this proposal can bring real benefits, consider a common smart manufacturing setting, such as a real-time quality control system on a production line.
The edge devices need to assess whether a product has a defect on time and catch all the defective ones.
It can base its decision on the inference provided by the early exit strategy, which is fairly accurate but not as accurate as the inference produced by the full \gls{dnn}.
As such, it will move the defective product to a buffer on the side of the line while the remote device runs the remaining layers of the \gls{dnn}.
Once the remote device completes the inference, a worker or a conveyor belt system can reintroduce the falsely marked non-defective products inside the production line.

In summary, the main contributions of this paper are:
\begin{itemize}
\item A strategy for implementing predefined sparsity, where a predefined set of sparse neuron connections is established before training and remains constant throughout training and inference.
\item Our approach involves the removal of specific connections within the neural network, reducing computational and storage complexity during inference and throughout the training process.
\item We apply this methodology in an \gls{sc} and \gls{ee} scenario, demonstrating its potential to further enhance application performance within this context.
\end{itemize}

The rest of the paper is organized as follows.
\cref{cha:related} presents background information.
Our proposal's details are outlined in \cref{cha:method}, followed by experiments in \cref{cha:experiments}.
Finally, conclusions are drawn in \cref{cha:conclusions}.

%% file: figures/front.tex
\begin{tikzpicture}[
    every node/.append style={
        align = center, font=\sffamily\small, outer sep=0pt, inner sep=2pt
    },
    neuron/.style={
        circle, draw, minimum size=0.25cm, line width=0.1, fill=yellow, opacity=.75, text opacity=1
    },
    connection/.style={draw, line width=0.5},
    box/.style={draw, thick, rounded corners=2pt, minimum width=2.20cm, minimum height=2.0cm},
    action/.style={draw, rounded corners=2pt, thick, minimum width=2.00cm, minimum height=1.00cm, fill=blue!30},
    device/.style={draw, rounded corners=2pt, thick, minimum width=2.00cm, minimum height=1.00cm, fill=orange!30},
]
\tikzmath{\distance=0.2;}

\node[box, minimum width=2.20cm, minimum height=2.20cm, AW]
    (dnn) at (0,0) {};
\node[AS, outer sep=2pt] at (dnn.south) {\textbf{DNN}};
\node[neuron, ANW] (n00) at ($(dnn.north west)+(\distance,-1.00*\distance)$) {};
\node[neuron, AN] (n01) at ($(n00.south)-(0,\distance)$) {};
\node[neuron, AN] (n02) at ($(n01.south)-(0,\distance)$) {};
\node[neuron, AN] (n03) at ($(n02.south)-(0,\distance)$) {};
\node[neuron, AW] (n10) at ($(n00)!.5!(n01)+(2*\distance,0)$) {};
\node[neuron, AW] (n11) at ($(n01)!.5!(n02)+(2*\distance,0)$) {};
\node[neuron, AW] (n12) at ($(n02)!.5!(n03)+(2*\distance,0)$) {};
\node[neuron, AW] (n20) at ($(n10)!.5!(n11)+(2*\distance,0)$) {};
\node[neuron, AW] (n21) at ($(n11)!.5!(n12)+(2*\distance,0)$) {};
\node[neuron, AW] (n30) at ($(n20)!.5!(n21)+(2*\distance,0)$) {};
\draw[connection] (n00.west -| dnn.west) -- (n00.west);
\draw[connection] (n01.west -| dnn.west) -- (n01.west);
\draw[connection] (n02.west -| dnn.west) -- (n02.west);
\draw[connection] (n03.west -| dnn.west) -- (n03.west);
\draw[connection] (n00.east) -- (n10.west);
\draw[connection] (n01.east) -- (n10.west);
\draw[connection] (n02.east) -- (n10.west);
\draw[connection] (n03.east) -- (n10.west);
\draw[connection] (n00.east) -- (n11.west);
\draw[connection] (n01.east) -- (n11.west);
\draw[connection] (n02.east) -- (n11.west);
\draw[connection] (n03.east) -- (n11.west);
\draw[connection] (n00.east) -- (n12.west);
\draw[connection] (n01.east) -- (n12.west);
\draw[connection] (n02.east) -- (n12.west);
\draw[connection] (n03.east) -- (n12.west);
\draw[connection] (n10.east) -- (n20.west);
\draw[connection] (n11.east) -- (n20.west);
\draw[connection] (n12.east) -- (n20.west);
\draw[connection] (n10.east) -- (n21.west);
\draw[connection] (n11.east) -- (n21.west);
\draw[connection] (n12.east) -- (n21.west);
\draw[connection] (n20.east) -- (n30.west);
\draw[connection] (n21.east) -- (n30.west);
\draw[connection] (n30.east) --++ (\distance,0);

\node[box, minimum width=6.50cm, minimum height=2.20cm, AW] (old)
    at ($(dnn.east)+(2*\distance,0)$) {};
\node[AN, outer sep=2pt] at (old.north) {\textbf{State-of-the-Art} approach};
\node[action, AW] (failing)
    at ($(dnn.east)+(4*\distance,0)$)
    {\textbf{Pruning} and\\\textbf{Quantization}};
\node[device, AW] (edge)
    at ($(failing.east)+(9*\distance,0)$) {\textbf{Edge}\\Device};

\draw[thick, ->] (dnn) -- (failing);
\draw[thick, ->] (failing) -- (edge)
    node[midway, font=\sffamily\footnotesize, AC, outer sep=2pt]
    {\emph{Training} and\\\emph{deploy}};

\node[box, minimum width=9.10cm, minimum height=3.40cm, ANW] (new)
    at ($(dnn.south west)+(0,-2*\distance)$) {};
    
\node[action, minimum width=1.75cm, AN] (sparsity)
    at ($(dnn.south)+(0,-8*\distance)$) {Apply\\\textbf{sparsity}};
\node[action, minimum width=1.5cm, AW] (ee)
    at ($(sparsity.east)+(2*\distance,0)$) {Apply\\\textbf{Split}};
\node[action, minimum width=1.5cm, AW] (training)
    at ($(ee.east)+(2*\distance,0)$) {Insert\\\textbf{Early Exit}};
\node[device, AS] (edge)
    at ($(training -| edge)+(0, +2*\distance)$) {\textbf{Edge}\\Device};
\node[device, AN] (remote)
    at ($(training -| edge)+(0, -2*\distance)$) {\textbf{Remote}\\Device};
\node[AW, font=\sffamily\footnotesize, outer sep=0.4cm]
    at (training.east) {\emph{Training} and\\\emph{deploy}};

\draw[thick, ->] (dnn) -- (sparsity);
\draw[thick, ->] (sparsity) -- (ee);
\draw[thick, ->] (ee) -- (training);
\draw[thick, ->] (training) -- (edge);
\draw[thick, ->] (training) -- (remote);

\end{tikzpicture}

%% file: figures/methodology.tex
\begin{tikzpicture}[
    every node/.append style={
        align = center, font=\sffamily\normalsize, outer sep=0pt, inner sep=0pt
    },
    neuron/.style={
        circle, draw, minimum size=0.5cm, line width=0.1, fill=yellow, opacity=.75, text opacity=1
    },
    EarlyExit/.style={neuron, densely dashed, fill=purple!75},
    Splitted/.style={neuron, fill=blue!75},
    connection/.style={draw, line width=0.5},
    box/.style={
        draw, ultra thick,
        rounded corners=4pt,
        minimum height=6.5cm,
    },
    Stage0/.style={box, minimum width=4.75cm, draw=blue!75},
	Stage1/.style={box, minimum width=5.50cm, draw=red!75},
	Stage2/.style={box, minimum width=6.00cm, draw=purple!75},
    properties/.style={align=left, font=\sffamily\small},
    cloud1/.append style={
        cloud, AS, draw =gray, cloud puffs = 8,
        text =black, fill = gray!40,
        minimum width = 1cm,
        minimum height = 1.25cm
    },
]
\tikzmath{\distance=0.3;}

\node[Stage0, AW] (Stage0) at (0,0) {};
\node[AN, outer sep=2pt] at (Stage0.north) {%
    Model $\mathcal{M}(\cdot{})$ that needs to\\%
    solve \textbf{real-world task}%
};
\node[AS, properties, outer sep=2pt] at (Stage0.south) {%
    \textcolor{red}{\tikzxmark} Computationally intensive\\
    \textcolor{red}{\tikzxmark} High storage requirements\\
    \textcolor{red}{\tikzxmark} High energy consumption
};
\node[neuron, ANW] (n00) at ($(Stage0)+(-4.5*\distance,4.5*\distance)$) {};
\node[neuron, AN] (n01) at ($(n00.south)-(0,\distance)$) {};
\node[neuron, AN] (n02) at ($(n01.south)-(0,\distance)$) {};
\node[neuron, AN] (n03) at ($(n02.south)-(0,\distance)$) {};
\node[neuron, AW] (n10) at ($(n00)!.5!(n01)+(2*\distance,0)$) {};
\node[neuron, AW] (n11) at ($(n01)!.5!(n02)+(2*\distance,0)$) {};
\node[neuron, AW] (n12) at ($(n02)!.5!(n03)+(2*\distance,0)$) {};
\node[neuron, AW] (n20) at ($(n10)!.5!(n11)+(2*\distance,0)$) {};
\node[neuron, AW] (n21) at ($(n11)!.5!(n12)+(2*\distance,0)$) {};
\node[neuron, AW] (n30) at ($(n20)!.5!(n21)+(2*\distance,0)$) {};
\draw[connection] (n00.west) --++ (-\distance, 0);
\draw[connection] (n01.west) --++ (-\distance, 0);
\draw[connection] (n02.west) --++ (-\distance, 0);
\draw[connection] (n03.west) --++ (-\distance, 0);
\draw[connection] (n00.east) -- (n10.west);
\draw[connection] (n01.east) -- (n10.west);
\draw[connection] (n02.east) -- (n10.west);
\draw[connection] (n03.east) -- (n10.west);
\draw[connection] (n00.east) -- (n11.west);
\draw[connection] (n01.east) -- (n11.west);
\draw[connection] (n02.east) -- (n11.west);
\draw[connection] (n03.east) -- (n11.west);
\draw[connection] (n00.east) -- (n12.west);
\draw[connection] (n01.east) -- (n12.west);
\draw[connection] (n02.east) -- (n12.west);
\draw[connection] (n03.east) -- (n12.west);
\draw[connection] (n10.east) -- (n20.west);
\draw[connection] (n11.east) -- (n20.west);
\draw[connection] (n12.east) -- (n20.west);
\draw[connection] (n10.east) -- (n21.west);
\draw[connection] (n11.east) -- (n21.west);
\draw[connection] (n12.east) -- (n21.west);
\draw[connection] (n20.east) -- (n30.west);
\draw[connection] (n21.east) -- (n30.west);
\draw[connection, ->] (n30.east)--++(+\distance,0);

\node[Stage1, AW] (Stage1) at ($(Stage0.east)+(2*\distance,0)$) {};
\node[AN, outer sep=2pt] at (Stage1.north)
    {Model $\mathcal{M}(\cdot{})$ after applying\\
    the \textbf{predefined sparsity}};
\node[AS, properties, outer sep=2pt] at (Stage1.south) {%
    \textcolor{green}{\tikzcmark} Not so computationally intensive\\
    \textcolor{green}{\tikzcmark} Less storage requirements\\
    \textcolor{green}{\tikzcmark} Less energy consumption
};
\node[neuron, ANW] (n00) at ($(Stage1)+(-4.5*\distance,4.5*\distance)$) {};
\node[neuron, AN] (n01) at ($(n00.south)-(0,\distance)$) {};
\node[neuron, AN] (n02) at ($(n01.south)-(0,\distance)$) {};
\node[neuron, AN] (n03) at ($(n02.south)-(0,\distance)$) {};
\node[neuron, AW] (n10) at ($(n00)!.5!(n01)+(2*\distance,0)$) {};
\node[neuron, AW] (n11) at ($(n01)!.5!(n02)+(2*\distance,0)$) {};
\node[neuron, AW] (n12) at ($(n02)!.5!(n03)+(2*\distance,0)$) {};
\node[neuron, AW] (n20) at ($(n10)!.5!(n11)+(2*\distance,0)$) {};
\node[neuron, AW] (n21) at ($(n11)!.5!(n12)+(2*\distance,0)$) {};
\node[neuron, AW] (n30) at ($(n20)!.5!(n21)+(2*\distance,0)$) {};
\draw[connection] (n00.west) --++ (-\distance, 0);
\draw[connection] (n01.west) --++ (-\distance, 0);
\draw[connection] (n02.west) --++ (-\distance, 0);
\draw[connection] (n03.west) --++ (-\distance, 0);
\draw[connection] (n02.east) -- (n10.west);
\draw[connection] (n03.east) -- (n10.west);
\draw[connection] (n01.east) -- (n11.west);
\draw[connection] (n00.east) -- (n12.west);
\draw[connection] (n10.east) -- (n20.west);
\draw[connection] (n12.east) -- (n20.west);
\draw[connection] (n11.east) -- (n21.west);
\draw[connection] (n20.east) -- (n30.west);
\draw[connection] (n21.east) -- (n30.west);
\draw[connection, ->] (n30.east)--++(+\distance,0);

\node[Stage2, AW] (Stage2) at ($(Stage1.east)+(2*\distance,0)$) {};
\node[AN, outer sep=2pt] at (Stage2.north)
    {Model $\mathcal{M}(\cdot{})$ after applying \textbf{Split}\\\
    \textbf{Computing} and \textbf{Early Exit}};
\node[AS, properties, outer sep=2pt] at (Stage2.south) {%
    \includegraphics[height=2ex]{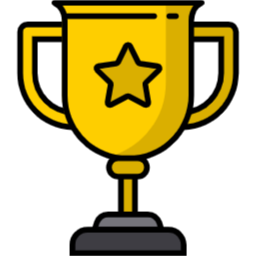}\textbf{State-of-the-art real-world task}\\
    \textcolor{green}{\tikzcmark} Not so computationally intensive\\
    \textcolor{green}{\tikzcmark} Less storage requirements\\
    \textcolor{green}{\tikzcmark} Less energy consumption
};

\tikzmath{\distance=0.2;}

\node[neuron, ANW] (n00) at ($(Stage2)+(-12*\distance,11*\distance)$) {};
\node[neuron, AN] (n01) at ($(n00.south)-(0,\distance)$) {};
\node[neuron, AN] (n02) at ($(n01.south)-(0,\distance)$) {};
\node[neuron, AN] (n03) at ($(n02.south)-(0,\distance)$) {};
\node[neuron, AW] (n10) at ($(n00)!.5!(n01)+(3*\distance,0)$) {};
\node[neuron, AW] (n11) at ($(n01)!.5!(n02)+(3*\distance,0)$) {};
\node[neuron, AW] (n12) at ($(n02)!.5!(n03)+(3*\distance,0)$) {};
\node[EarlyExit, AW] (ee)  at ($(n12)+(4*\distance,-1*\distance)$) {};
\node[Splitted, AW] (n20) at ($(n10)!.5!(n11)+(12*\distance,0)$) {};
\node[Splitted, AW] (n21) at ($(n11)!.5!(n12)+(12*\distance,0)$) {};
\node[Splitted, AW] (n30) at ($(n20)!.5!(n21)+(4*\distance,0)$) {};
\node[cloud1] (net) at ($(ee.north)+(2*\distance,1*\distance)$){};
\node[AS, outer sep=1pt] at (net.north) {\emph{Network}};
\draw[connection] (n00.west) --++ (-\distance, 0);
\draw[connection] (n01.west) --++ (-\distance, 0);
\draw[connection] (n02.west) --++ (-\distance, 0);
\draw[connection] (n03.west) --++ (-\distance, 0);
\draw[connection] (n02.east) -- (n10.west);
\draw[connection] (n03.east) -- (n10.west);
\draw[connection] (n01.east) -- (n11.west);
\draw[connection] (n00.east) -- (n12.west);
\draw[mark=none, black, dotted, line width=1.5] (n10.east) -- (n20.west);
\draw[mark=none, black, dotted, line width=1.5] (n12.east) -- (n20.west);
\draw[mark=none, black, dotted, line width=1.5] (n11.east) -- (n21.west);
\draw[connection, densely dashed] (n10.east) -- (ee.west);
\draw[connection, densely dashed] (n12.east) -- (ee.west);
\draw[connection, densely dashed] (n11.east) -- (ee.west);
\draw[connection, densely dashed, ->] (ee.east)--++(+2*\distance,0);
\draw[connection] (n20.east) -- (n30.west);
\draw[connection] (n21.east) -- (n30.west);
\draw[connection, ->] (n30.east)--++(+\distance,0);
\draw[decorate, thick, decoration={brace, amplitude=3pt, mirror, raise=0.1cm}]
  ($(n03.south west)+(-1.5*\distance,0)$) --
  ($(n03.south west -| n12.south east)+(1.5*\distance,0)$)
  node[midway,yshift=-0.75cm]{%
  $\mathcal{M}(\cdot{})$ head\\%
  on \textbf{Edge}%
};
\draw[decorate, thick, decoration={brace, amplitude=3pt, mirror, raise=0.1cm}]
  ($(n03.south west -| ee.south west)+(-0.50*\distance,0)$) --
  ($(n03.south west -| ee.south east)+(+2.25*\distance,0)$)
  node[midway,yshift=-0.75cm]{%
  \textbf{Early}\\\textbf{exit}%
};
\draw[decorate, thick, decoration={brace, amplitude=3pt, mirror, raise=0.1cm}]
  ($(n03.south west -| n21.south west)+(-1*\distance,0)$) --
  ($(n03.south west -| n30.south east)+(+1*\distance,0)$)
  node[midway,yshift=-0.75cm]{%
  $\mathcal{M}(\cdot{})$ tail\\%
  on \textbf{Remote}%
};

\draw[ultra thick, ->] (Stage0) -- (Stage1);
\draw[ultra thick, ->] (Stage1) -- (Stage2);

\end{tikzpicture}

%% file: srcs/2_related.tex
\section{Related Work}
\label{cha:related}

This section provides an overview of \textit{i)} \glspl{dnn} sparsity methods, and \textit{ii)} distributed deep learning applications with a specific focus on \gls{sc} and \gls{ee}.

\subsection{Deep Neural Networks (DNNs) sparsity}
While numerous methods concerning sparse \glspl{dnn} can be found in the literature~\cite{gong2014compressing,chen2015compressing,wen2016learning}, most of these do not reduce the computation and storage complexity associated with training, but only with inference.

An example is the dropout.
This technique is used to prevent overfitting.
It randomly drops some neurons out (\ie{}, ignoring) during each training iteration.
This helps the network learn more robust and generalized features, as it can't rely too much on any one neuron~\cite{srivastava2014dropout}.
Due to lack of space, and since it is only a related topic, refer to~\cite{labach2019survey} for more details.

There are other approaches, such as pruning and trimming methods, that process the trained \gls{dnn} to generate a sparse \gls{dnn} for inference.
For example, in~\cite{albericio2016cnvlutin}, the authors observe that a large fraction of the computations performed by \glspl{dnn} are intrinsically ineffectual as they involve a multiplication where one of the inputs is zero.
This observation motivates ``Cnvlutin'', a value-based approach that eliminates most of the ineffectual operations.
Furthermore, in~\cite{aghasi2017net}, the authors present an algorithm that prunes (specifies) a trained network layer-wise, removing connections at each layer by solving a convex optimization program.
This program seeks a sparse set of weights at each layer, keeping the layer inputs and outputs consistent with the originally trained model. 

As mentioned in Section~\ref{cha:intro}, other methods aim to reduce the complexity of performing inference on a trained \gls{dnn}.
These methods encompass techniques like quantization~\cite{gong2014compressing} and compression~\cite{han2015deep}.
It's important to note that all these methods primarily target reducing complexity in the inference models rather than significantly simplifying the training process.

One approach that strives to minimize complexity in both training and inference involves using neural networks with structured weight matrices that are not necessarily sparse~\cite{sindhwani2015structured}.

Lastly, it's worth mentioning that several authors have recently introduced the concept of predefined sparse neural networks~\cite{dey2019pre}. 
This surge in research activity is motivated by the recognition that specialized hardware is typically required to run large and complex neural networks effectively.

In this article, we leverage the previously mentioned \emph{predefined sparsity}.
Specifically, our approach offers two key benefits.
Firstly, it is hardware agnostic.
This is in contrast to quantization, which relies on specific hardware capabilities.
Secondly, it allows us to streamline our model before the training begins.
This differs from pruning, which tackles complexity reduction after training.
As a result, our model has a smaller footprint on the \gls{gpu} right from the start of training.

Our work goes a step further.
While default sparsity has been explored previously, we are the first to investigate its advantages within the context of \gls{sc} and \gls{ee}.
This exploration aims to unlock even greater efficiency for applications that utilize this technique.

\subsection{Distributed deep learning}
We focus on architectures operating through a \gls{dnn} model $\mathcal{M}(\cdot{})$, whose task is to produce the inference output $y$ from an input $x$.
We can identify three major types of architectures used for distributed deep learning applications in the literature: \gls{loc}, \gls{roc}, and \gls{sc}.

\paragraph*{\textbf{Local-only Computing (\acrshort{loc})}}
Under this policy, the entire computation is performed on the sensing devices.
Therefore, the edge device entirely executes the function $\mathcal{M}(x)$.
Its advantage lies in offering low latency due to the proximity of the computing element to the sensor~\cite{capogrosso2024machine}.
However, it may not be compatible with \gls{dnn}-based architectures that demand robust hardware capabilities.
Usually, simpler \gls{dnn} models $\bar{\mathcal{M}}(x)$ that use specific architectures (\eg{}, depth-wise separable convolutions) are used to build lightweight networks, such as MobileNetV3~\cite{howard2019searching}.

Besides designing lightweight neural models, in the last few years, great progress has been made in the area of \gls{dnn} compression.
Compression techniques, such as network pruning and quantization~\cite{liang2021pruning}, or knowledge distillation~\cite{gou2021knowledge} achieve a more efficient representation of one or more layers of the neural network, but with a possible quality degradation.

\paragraph*{\textbf{Remote-only Computing (\acrshort{roc})}}
The input $x$ is transferred through the communication network and then is processed at the remote system through the function $\mathcal{M}(x)$.
This architecture preserves full accuracy considering the higher power budget of the remote system, but it leads to high latency and bandwidth consumption due to the input transfer.

\paragraph*{\textbf{Split Computing (\acrshort{sc})}}
A typical \gls{sc} scenario is discussed in~\cite{eshratifar2019jointdnn}, where the authors show that neither \gls{loc} nor \gls{roc} approaches are optimal, and a split configuration is an ideal solution.
The \gls{sc} paradigm divides the \gls{dnn} model into a head, executed by the local sensing device, and a tail, executed by the remote system.
It combines the advantages of both \gls{loc} and \gls{roc} thanks to the lower latency and, more importantly, drastically reduces the required transmission bandwidth by compressing the input to be sent $x$ through the use of an autoencoder~\cite{matsubara2019distilled}.
We define the encoder and decoder models as $z_{l} = \mathcal{F}(x)$ and $\bar{x} = \mathcal{G}(z_{l})$, which are executed at the edge, and remotely, respectively.
The distance $d(x,\bar{x})$ defines the performance of the encoding-decoding process.

One of the earliest works on \gls{sc} is the study by Kang~\etal{}~\cite{kang2017neurosurgeon}, in which the authors show that the initial layers of a \gls{dnn} are the most suitable candidates for partitioning, as they optimize both latency and energy consumption.
Additionally, latency reduction is usually achieved through quantization, as explored in~\cite{li2018auto}, and the utilization of lossy compression techniques prior to data transmission, as investigated in~\cite{choi2018deep}.
In addition to lossy compression techniques, in~\cite{carra2023dnn}, the authors also explore lossless techniques to encode intermediate results without modifying the \gls{ml} model.
Instead, the concept of employing autoencoders to compress the data further to be transferred is discussed in various studies, such as~\cite{eshratifar2019bottlenet}.

The prevalent methods for identifying potential splitting points have evolved from architecture-based techniques to more refined neuron-based methods.
Within the domain of architecture-based approaches, in~\cite{sbai2021cut}, the authors state that candidate split locations are where the size of the \gls{dnn} layers decreases: the rationale is that compressing information by autoencoders, where compression would still occur due to the shrinking of the architecture, certainly seems reasonable.
In~\cite{cunico2022split} and~\cite{capogrosso2023split}, the authors show that not only the architecture of the layers but also the saliency of individual layers is a crucial factor when deciding where to split.
A neuron's saliency is determined by its gradient in relation to the accurate decision.
Consequently, optimal splitting points should be strategically positioned following layers housing a concentration of impactful neurons to preserve the information flowing until then.

At the same time, current state-of-the-art approaches in different \gls{ml} applications rely on advanced learning procedures, such as the \emph{\gls{mtl}}~\cite{caruana1997multitask}.
In particular, \gls{mtl} is a paradigm in which multiple related tasks are jointly learned to improve the generalizability of a model by using shared knowledge across different aspects of the input.
As a result, in~\cite{capogrosso2024mtl}, the authors propose, for the first time ever, how to partition multi-tasking \gls{dnn} to be deployed within a \gls{sc} framework.
With this design, the authors can handle multiple tasks concurrently instead of the current focus on \gls{stl} in \gls{sc}, and through \gls{mtl}, they increase task performance, overcoming the challenge of preserving only the performance of the main task.

\paragraph*{\textbf{Early Exit (\acrshort{ee})}}
This scenario adds an early exiting branch to a standard \gls{sc} architecture.
Formally, we can define $B_{i},i=1\dots{}N$ (with $N=\layers$, and $\layers$ is the number of layers of the \gls{dnn}) as the branch model that takes as input $z_{l}$ and produces an estimate of the desired output $y$.
In practice, the \gls{ee} architecture is a modification of an existing neural network, adding one or more classification branches where, before the computation of all network's layers, the confidence of the intermediate result is checked to be enough to be considered the final result~\cite{lo2017dynamic}. 

\gls{ee} architecture can be exploited in a distributed deep learning application where the intermediate result can be directly transmitted, as in local computing, or refined at the remote side, as in \gls{sc}.
In this scenario, the level of transmission traffic depends on the input, thus varying stochastically.
Therefore, the interdependencies between computation and communication cannot be analytically modeled, and real experiments are needed to validate a given implementation.

In this paper, we're not looking to develop a new method for finding the best layers in a \gls{dnn} to use \gls{sc} or the most efficient way to put \gls{ee} into action.
We aim to explore and show how the predefined sparsity can improve these existing frameworks.
Thus, instead of comparing our work to \gls{sc} and \gls{ee} methods, we see it as a way to make them even better, investigating how sparsity can add benefits on top of what \gls{sc} and \gls{ee} already offer.

%% file: srcs/3_method.tex
\section{Method} 
\label{cha:method}

To understand the core concepts of our research, let us delve into the mathematical background.
These concepts were initially presented in~\cite{dey2019pre}.

Let us take a $(\layers + 1)$-layer \gls{mlp}, described collectively by the following \textit{neuronal configuration}:
\begin{equation*}
    \neurons_{net} = (\neurons_{0}, \dots, \neurons_{i-1}, \neurons_{i}, \dots, \neurons_{\layers})\mcomma
\end{equation*}
where $\neurons_{i}$ represents the number of nodes in the $i$-th layer.
We use the convention that layer $i$ is to the right of layer $i-1$.
Given $\layers$ \textit{junctions} between layers, with junction $i$ connecting the $\neurons_{i-1}$ nodes of its left layer ${i-1}$ with the $\neurons_{i}$ nodes of its right layer $i$.

We can define \textit{predefined sparsity} as simply not having all $\neurons_{i-1} \cdot \neurons_{i}$ edges (or weights) present in junction $i$.
Furthermore, we can define \textit{structured predefined sparsity} so that for a given junction $i$, each node in its left layer has fixed out-degree, \ie{}, $\outdegree_{i}\in\mathbb{N}$ connections to its right layer, and each node in its right layer has fixed in-degree, \ie{}, $\indegree_{i}\in\mathbb{N}$ connections from its left layer.

In particular, a \gls{mlp} have $\outdegree_{i} = \neurons_{i}$ and $\indegree_{i} = \neurons_{i-1}$ with $\neurons_{i-1} \cdot \neurons_{i}$ edges present in the $i^{th}$ junction.
While a sparse \gls{mlp} has at least one junction with less than this number of edges.
The number of edges in junction $i$ is given by the formula:
\begin{equation}\label{eq:edges}
    \edges{i} = \neurons_{i-1} \cdot \outdegree_{i} = \neurons_{i} \cdot \indegree_{i}\mperiod
\end{equation}
The density of junction $i$ is measured relative to \gls{mlp}, and is denoted by the function:
\begin{equation}\label{eq:density}
    \density_{i} = \frac{\edges{i}}{\neurons_{i-1} \cdot \neurons_{i}}\mperiod
\end{equation}

In our structured predefined sparse network, the density of the $i$-th junction $\density_{i}$ cannot be arbitrary.
By replacing \cref{eq:edges} in \cref{eq:density}, we can define:
\begin{equation}
    \density_{i}
    = \frac{\indegree_{i}}{\neurons_{i}}
    = \frac{\neurons_{i}}{\outdegree_{i}}\mcomma
\end{equation}
where $\outdegree_{i}$ and $\indegree_{i}$ are natural numbers.
The number of possible $\density_{i}$ values is the same as the number of ($\outdegree_{i}$, $\indegree_{i}$) values satisfying the structured predefined sparsity constraints:
\begin{equation}\label{eq:constraints}
    \outdegree_{i} = \frac{\neurons_{i} \cdot \indegree_{i}}{\neurons_{i-1}},
    \qquad
    \indegree_{i}\leq{}\neurons_{i-1}\;\mperiod
\end{equation}

Now, the smallest value of $\indegree_{i}$ which satisfies the assignment to $\outdegree_{i}$ in \cref{eq:constraints}, and $\outdegree_{i}\in\mathbb{N}$, is the following:
\begin{equation*}
    \frac{\neurons_{i-1}}{\gcd(\neurons_{i-1},\neurons_{i})}\mcomma
\end{equation*}
and other values are integer multiples.
Since $\indegree_{i}$ is upper bounded by $\neurons_{i-1}$, the total number of possible ($\outdegree_{i}, \indegree_{i}$) is $\gcd(\neurons_{i-1}, \neurons_{i})$.
We can now define the set of possible densities $\density_{i}$, as follows:
\begin{equation}
    \Biggl\{
    \density_{i}
    = \frac{k}{\gcd(\neurons_{i-1},\neurons_{i})},
    \quad
    \density_{i}\in{}(0,1],
    \quad
    k\in{}\mathbb{N}\Biggr\}\;.
\end{equation}

Specifying $\neurons_{net}$ and the \textit{out-degree configuration} $\outdegree_{net}=(\outdegree_{1},\dots{},\outdegree_{\layers})$ determines the density of each junction and the overall density, defined as:
\begin{equation}
    \density_{net}
    = \frac{\sum_{i=1}^{\layers}{\edges{i}}}
           {\sum_{i=1}^{\layers}{\neurons_{i-1}\cdot\neurons_{i}}}\;.
\end{equation}

It's worth noting that, for an \gls{mlp} using structured predefined sparsity, only the weights corresponding to connected edges are stored in memory and used in the computation.
Specifically, the parameters are updated based on the gradient of a loss function with respect to the parameters.
Let's denote the parameters of the network as $\theta{}$ and the loss function as $\mathcal{\layers}(\theta{})$.
The gradient of the loss function for the parameters is denoted as $\nabla_{\theta{}}\mathcal{\layers}$.
Then, the parameters are updated using a gradient descent step:
\begin{equation}
    \theta_{\text{new}} =
    \theta_{\text{old}} - \eta \cdot \nabla_{\theta}\mathcal{\layers}\mcomma
\end{equation}
where $\eta$ is the learning rate, controlling the step size in the direction opposite to the gradient.

%% file: srcs/4_experiments.tex
\section{Experiments}
\label{cha:experiments}

This section describes the experimental trials performed to validate our proposal, along with their implementation details and results.

\begin{figure*}[!ht]
    \centering
    \subfigure[Weights distribution in layer 1 {[800, 180]}.]
    {\input{graphs/hist_layer1}\label{fig:hist_layer1}}%
    \subfigure[Weights distribution in layer 2 {[180, 180]}.]
    {\input{graphs/hist_layer2}\label{fig:hist_layer2}}%
    \subfigure[Weights distribution in layer 3 {[800, 10]}.]
    {\input{graphs/hist_layer3}\label{fig:hist_layer3}}%
    \caption{Histograms of weights in each junction resulting from training a deep \gls{mlp} on the MNIST dataset.
    The network configuration $H=[H_{0},\dots{},H_{n}]$ used is [800, 180, 180, 10].
    }
    \label{fig:pdsexample}
\end{figure*}
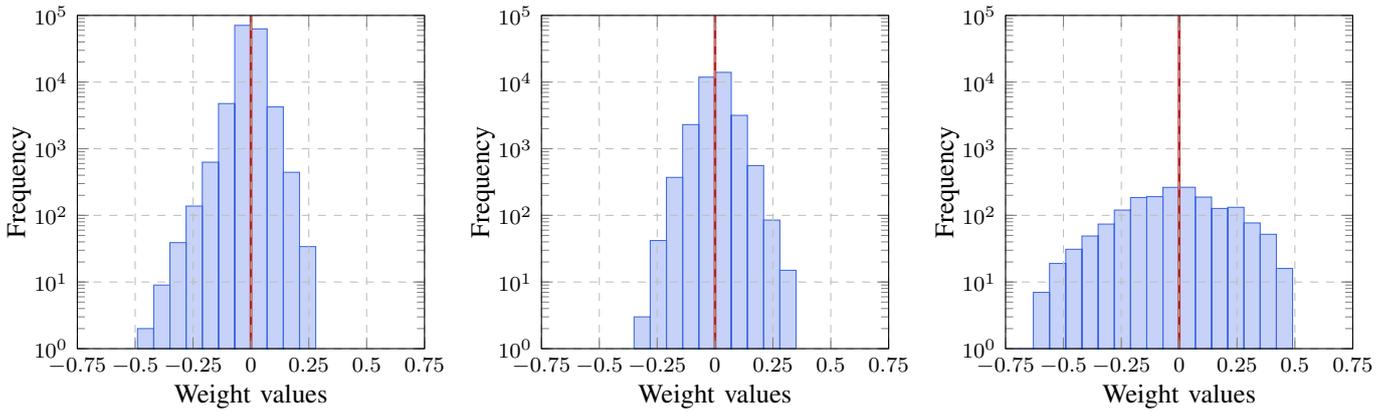

\paragraph*{\textbf{Models details}}
We use three types of \glspl{mlp} in our experiments: shallow, deep, and sparse.
We can characterize them as follows:
\begin{itemize}
\item \textbf{Shallow}: This is the simplest type.
It contains only one hidden layer besides the input and output layers and represents the base case.
The hidden layer neurons receive information from the input layer and process it before transmitting it to the final output layer;
\item \textbf{Deep}: Each neuron within a hidden layer is fully connected to all neurons in the subsequent layer;
\item \textbf{Sparse}: The \emph{predefined sparsity} pattern is applied.
This sparsity pattern essentially removes certain connections between neurons in adjacent layers.
Specifically, the sparse \gls{mlp} aims to balance the simplicity of the shallow \gls{mlp} and the learning capacity of the deep \gls{mlp}.
\end{itemize}

Each network configuration is defined by using two lists.
The first list identifies the number of neurons in the hidden layers:
\begin{equation*}
    H=[H_{0}, \dots, H_{n}]\mcomma
\end{equation*}
where $n$ is the number of hidden layers, and intuitively, $H_{0}$ and $H_{n}$ represent the number of neurons in the input and output layers, respectively.
The second list identifies the out-degree of each neuron for each layer:
\begin{equation*}
    G=[G_{0}, \dots, G_{L - 1}]\mperiod
\end{equation*}
Specifically, $G_{0}$ is the out-degree of each input layer node, and the output layer $G_{L}$ is not reported because it is zero.

Then, the shallow, deep, and sparse \gls{mlp} has been split.
Our research primarily focuses on using existing sparsity patterns in split computing applications for these networks.
Inspired by the approach presented in~\cite{kang2017neurosurgeon}, we opt to split the network at the midpoint, ensuring uniformity in our approach.
While more recent studies, such as~\cite{sbai2021cut,cunico2022split}, have proposed advanced techniques for selecting optimal splitting points, exploring these methods will be part of future investigations.

Finally, the inserted \glspl{ee} are composed of linear layers followed by \gls{relu} activation functions, where the dimension of the last layer matches the desired output size of the \gls{dnn}.

Even with the rise of specialized architectures like \glspl{cnn}, \glspl{mlp} remains a valuable tool for researchers.
Their fundamental structure allows for in-depth exploration of core deep learning concepts without the added complexity of specialized architectures.
This focus on \glspl{mlp} aligns with the experimental nature of this research, where we aim to isolate the effects of the proposed technique and gain a deeper understanding of its fundamental workings.
Furthermore, recent advancements in research in \gls{mlp} have demonstrated their continued effectiveness in various real-world applications, highlighting their ongoing relevance in the field~\cite{turetta2023towards,chen2023tsmixer}.

\paragraph*{\textbf{Datasets}}
In this research, we focus on the image classification task.
We utilize the MNIST dataset~\cite{deng2012mnist}, a well-known collection of handwritten digits.
This contains 60,000 training and 10,000 testing images for the multi-class image classification task.
The images were centered in a $28\times{}28$ image by computing the center of mass of the pixels and translating the image to position this point at the center of the $28\times{}28$ field.
 
MNIST has to be considered as a placeholder for bigger datasets (\eg{}, ImageNet~\cite{deng2009imagenet}); nonetheless, the focus here is to show the potentialities of the predefined sparsity applied in an \gls{sc} and \gls{ee} and not beating the state-of-the-art in a specific computer vision challenge.

While this article focuses on a computer vision task, the concepts explored here remain valid also for broader applications.
For example, they can be effectively used for other tasks, such as time series forecasting.

\paragraph*{\textbf{Training details}}
All the source code is implemented in TensorFlow~\cite{tensorflow2015}.
We train our models for 50 epochs, with a learning rate of $1\times{}10^{-5}$, using Adam~\cite{kingma2014adam} as an optimizer, on an NVIDIA RTX 3060 Ti.

\subsection{Why predefined sparsity in \glsentryname{sc} and \glsentryname{ee}?}
The motivation behind predefined sparsity can be exemplified by examining the weights histograms of a trained \gls{mlp} shown in \cref{fig:pdsexample}.
Specifically, the three histograms show the weight distributions for each layer in a 3-layer \gls{mlp} model trained on the MNIST dataset with hidden layers having the following number of neurons $H=[800, 180, 180, 10]$.

As we can see from \cref{fig:hist_layer1,fig:hist_layer2}, the first layers of the network have a significant concentration of weights around zero, suggesting that these weights might not be crucial for the network's performance.
While \cref{fig:hist_layer3} highlights how the weights in the last layer assume a broader spectrum of values.

This finding suggests that the benefits of \emph{predefined sparsity} can be especially pronounced in the earlier layers of the network.
As a result, in resource-constrained environments, like those encountered in \gls{sc} and \gls{ee} applications, predefined sparsity offers significant advantages because by reducing the number of connections, we can decrease the size and complexity of the network portion deployed on the edge device. 
This translates to lower memory requirements and faster processing times during training and inference.

Furthermore, the storage footprint is directly proportional to the number of edges.
Operating at a sparsity level of, for example, 50\% results in a two-fold reduction in complexity.
This translates to significant efficiency gains, making deploying more complex models on devices with limited resources possible.
These findings led us to want to study them in the context of \gls{sc} and \gls{ee}.

\subsection{Results}
\begin{table}[!t]
    \centering
    \caption{Results regarding the accuracy and the number of parameters with different configurations of head \glspl{mlp}.}
    \renewcommand{\arraystretch}{1.3}
    \input{tables/exp_head}
    \label{tab:exp_head}
\end{table}
\begin{table}[!t]
    \centering
    \caption{Results regarding the accuracy and the number of parameters with different configurations of tail \glspl{mlp}.}
    \renewcommand{\arraystretch}{1.3}
    \input{tables/exp_tail}
    \label{tab:exp_tail}
\end{table}

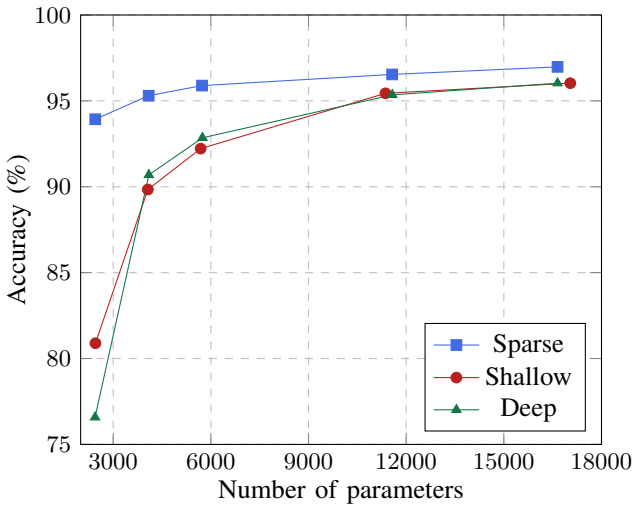
\begin{figure}[!t]
    \centering
    \input{graphs/accuracy_head}
    \caption{Accuracy tests by the number of parameters of deep, shallow, and sparse head \glspl{mlp}.}
    \label{fig:accuracy_head}
\end{figure}
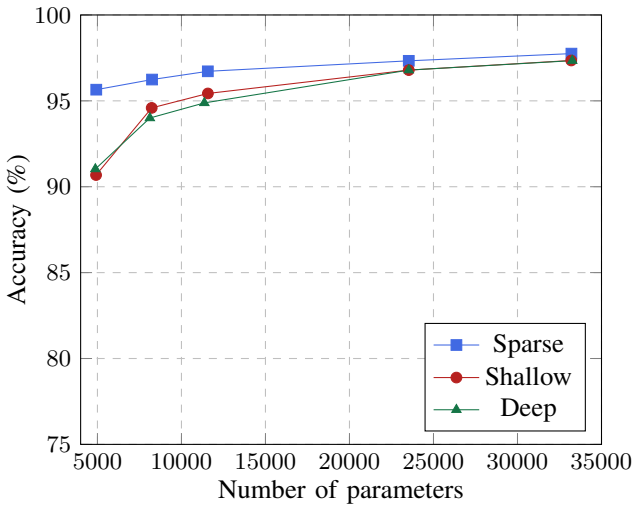
\begin{figure}[!t]
    \centering
   \input{graphs/accuracy_tail}
    \caption{Accuracy tests by number of parameters of deep, shallow, and sparse tail \glspl{mlp}.}
    \label{fig:accuracy_tail}
\end{figure}

\cref{tab:exp_head,tab:exp_tail} provide a comparative overview of the three neural network configurations, detailing their architectural details and performances.
Specifically, each table reports the number of neurons per layer, focusing on the number of neurons in hidden layers (in bold), the out-degree for each node, their accuracy, and the number of parameters.

\cref{fig:accuracy_head,fig:accuracy_tail} show the accuracy plots against the number of parameters for the three neural network configurations.
These two plots reveal the advantage of sparse split models: they present remarkable stability in accuracy even with significant reductions in trainable parameters.
This characteristic results in two key advantages, particularly desired and pursued in resource-constrained settings like \gls{sc} and \gls{ee}.

Unlike traditional deep and shallow \gls{dnn}, where accuracy improvements often depend on a significant increase in the number of parameters, sparse split models achieve optimal inference performance without requiring massive parameter expansions.
As a result, they are much more efficient when using the limited processing power and storage space available on edge devices, and so very suitable for distributed deep leasing scenarios through \gls{sc} and \gls{ee}.

Furthermore, the predefined sparsity of these networks leads to decreased memory usage, both during training and inference.
Regarding the training, this is an advantage because researchers with limited resources can also engage in cutting-edge research in deep learning without the need for expensive, high-end computational resources.
Instead, the reduced complexity translates into faster processing times, enabling real-time operation on edge devices with limited processing power.

\subsection{Discussion}

\paragraph*{\textbf{Early Exiting and reduced communication}}
In \gls{sc} and \gls{ee} applications, where data security and bandwidth limitations are paramount, sparse split models represent an ideal solution.

The early exiting strategy applied to the head provides an acceptable level of accuracy with a smaller subset of parameters, as highlighted by the achieved accuracy shown in \cref{fig:accuracy_head}.
This allows the network to terminate computations early, significantly reducing the amount of data transmitted between the edge device and the server.
This not only minimizes the risk of data breaches but also conserves precious network bandwidth, enabling efficient communication even in low-connectivity environments.

\paragraph*{\textbf{Balancing training time and efficiency}}
While the benefits of sparse split models are evident, it's important to acknowledge the trade-off with training time.
When we introduce predefined sparsity into a \glspl{mlp}, we aim to train the model to learn a function using a significantly reduced number of connections compared to a non-sparse network.
This approach creates a more lightweight structure for the model but also leads to a more challenging optimization problem.
Consequently, the gradient descent algorithm may require more iterations to converge due to the increased complexity introduced by the sparsity constraints.
In particular, sparse split models in our experiments exhibited a $2\times{}$ slowdown in training time compared to their dense counterparts. 

However, this drawback needs to be considered in the context of the specific application and its resource constraints.
In many cases, the significant gains in memory usage, communication efficiency, and inference speed during deployment far outweigh the potential increase in training time.

\paragraph*{\textbf{Potential further exploration}}
While this work demonstrates the effectiveness of predefined sparsity with \gls{sc} and \gls{ee}, further exploration is possible.
First, future research could look at applying this technique to different \gls{dnn} architectures on top of existing \gls{sc} and \gls{ee} methods.

It would also be interesting to see how different hardware edge devices can benefit from this.
For example, we know that devices like \glspl{gpu} and \glspl{fpga} handle sparse data well.
So, using this technique with these devices on the edge (like \gls{gpu}-based or \gls{fpga}-based hardware) could be a promising direction.

Finally, further exploration could involve applying these techniques to real-world applications.
This would allow us to assess the effectiveness of the approach in practical scenarios and identify any challenges that may arise.

%% file: graphs/hist_layer1.tex
\begin{tikzpicture}
\begin{axis}[
    height=6cm, width=6.15cm,
    area style,
    bar width=0.5,
    xmin=-0.75,
    xmax=+0.75,
    ymin=1,
    ymax=1e5,
    ymode=log,
    enlarge y limits={0},
    enlarge x limits={0},
    xtick distance={0.25},
    tick label style={font=\footnotesize},
    xlabel={Weight values},
    ylabel={Frequency},
    xmajorgrids=true,
    ymajorgrids=true,
    grid style=dashed,
    xlabel style={yshift=+0.1cm},
    ylabel style={yshift=-0.2cm},
]
\addplot+[ybar interval,mark=no] plot coordinates {%
    (-0.70, 0)
    (-0.63, 0)
    (-0.56, 0)
    (-0.49, 2)
    (-0.42, 9)
    (-0.35, 39)
    (-0.28, 138)
    (-0.21, 627)
    (-0.14, 4758)
    (-0.07, 70947)
    ( 0.00, 62759)
    ( 0.07, 4243)
    ( 0.14, 443)
    ( 0.21, 34)
    ( 0.28, 1)
    ( 0.35, 0)
    ( 0.42, 0)
    ( 0.49, 0)
    ( 0.56, 0)
    ( 0.63, 0)
    ( 0.70, 0)
};
\draw[line width=1pt, red] (0,1) -- (0, 1e05);
\end{axis}
\end{tikzpicture}

%% file: graphs/hist_layer2.tex
\begin{tikzpicture}
\begin{axis}[
    height=6cm, width=6.15cm,
    area style,
    bar width=0.5,
    xmin=-0.75,
    xmax=+0.75,
    ymin=1,
    ymax=1e5,
    ymode=log,
    enlarge y limits={0},
    enlarge x limits={0},
    xtick distance={0.25},
    tick label style={font=\footnotesize},
    xlabel={Weight values},
    ylabel={Frequency},
    xmajorgrids=true,
    ymajorgrids=true,
    grid style=dashed,
    xlabel style={yshift=+0.1cm},
    ylabel style={yshift=-0.2cm},
]
\addplot+[ybar interval,mark=no] plot coordinates {%
    (-0.70, 0)
    (-0.63, 0)
    (-0.56, 0)
    (-0.49, 0)
    (-0.42, 0)
    (-0.35, 3)
    (-0.28, 42)
    (-0.21, 371)
    (-0.14, 2302)
    (-0.07, 11859)
    ( 0.00, 13996)
    ( 0.07, 3166)
    ( 0.14, 558)
    ( 0.21, 85)
    ( 0.28, 15)
    ( 0.35, 3)
    ( 0.42, 0)
    ( 0.49, 0)
    ( 0.56, 0)
    ( 0.63, 0)
    ( 0.70, 0)
};
\draw[line width=1pt, red] (0,1) -- (0, 1e05);
\end{axis}
\end{tikzpicture}

%% file: graphs/hist_layer3.tex
\begin{tikzpicture}
\begin{axis}[
    height=6cm, width=6.15cm,
    area style,
    bar width=0.5,
    xmin=-0.75,
    xmax=+0.75,
    ymin=1,
    ymax=1e5,
    ymode=log,
    enlarge y limits={0},
    enlarge x limits={0},
    xtick distance={0.25},
    tick label style={font=\footnotesize},
    xlabel={Weight values},
    ylabel={Frequency},
    xmajorgrids=true,
    ymajorgrids=true,
    grid style=dashed,
    xlabel style={yshift=+0.1cm},
    ylabel style={yshift=-0.2cm},
]
\addplot+[ybar interval,mark=no] plot coordinates {%
    (-0.70, 1)
    (-0.63, 7)
    (-0.56, 19)
    (-0.49, 31)
    (-0.42, 49)
    (-0.35, 74)
    (-0.28, 120)
    (-0.21, 185)
    (-0.14, 191)
    (-0.07, 263)
    ( 0.00, 264)
    ( 0.07, 188)
    ( 0.14, 127)
    ( 0.21, 132)
    ( 0.28, 77)
    ( 0.35, 52)
    ( 0.42, 16)
    ( 0.49, 4)
    ( 0.56, 0)
    ( 0.63, 0)
    ( 0.70, 0)
};
\draw[line width=1pt, red] (0,1) -- (0, 1e05);
\end{axis}
\end{tikzpicture}

%% file: tables/exp_head.tex
\begin{tabular}{ll@{ }c@{ }rrcc}
\toprule
Type &
\multicolumn{3}{@{}c@{}}{\begin{tabular}{@{}c@{}}Neurons\\per Layer\end{tabular}} &
\begin{tabular}{@{}c@{}}Out-degree\\per Node\end{tabular} &
\begin{tabular}{@{}c@{}}Accuracy\\(\%)\end{tabular} &
\begin{tabular}{@{}c@{}}Number of\\Parameters\end{tabular} \\
\midrule
Deep    & [800 & \bfseries 3 3   & 10] & [3 3 10]   & 76.58 &  2443 \\
Shallow & [800 & \bfseries 3     & 10] & [3 10]     & 80.89 &  2455 \\
Sparse  & [800 & \bfseries 40 40 & 10] & [2 9 10]   & 93.93 &  2450 \\\hline
Deep    & [800 & \bfseries 5 5   & 10] & [5 5 10]   & 90.69 &  4095 \\
Shallow & [800 & \bfseries 5     & 10] & [5 10]     & 89.84 &  4065 \\
Sparse  & [800 & \bfseries 40 40 & 10] & [4 10 10]  & 95.30 &  4090 \\\hline
Deep    & [800 & \bfseries 7 7   & 10] & [7 7 10]   & 92.85 &  5743 \\
Shallow & [800 & \bfseries 7     & 10] & [7 10]     & 92.22 &  5687 \\
Sparse  & [800 & \bfseries 40 40 & 10] & [6 11 10]  & 95.89 &  5730 \\\hline
Deep    & [800 & \bfseries 14 14 & 10] & [14 14 10] & 95.35 & 11574 \\
Shallow & [800 & \bfseries 14    & 10] & [14 10]    & 95.44 & 11364 \\
Sparse  & [800 & \bfseries 40 40 & 10] & [13 17 10] & 96.54 & 11570 \\\hline
Deep    & [800 & \bfseries 20 20 & 10] & [20 20 10] & 96.03 & 16650 \\
Shallow & [800 & \bfseries 21    & 10] & [21 10]    & 96.03 & 17041 \\
Sparse  & [800 & \bfseries 40 40 & 10] & [19 24 10] & 96.98 & 16650 \\
\bottomrule
\end{tabular}%

%% file: tables/exp_tail.tex
\begin{tabular}{ll@{ }c@{ }rrcc}
\toprule
Type &
\multicolumn{3}{@{}c@{}}{\begin{tabular}{@{}c@{}}Neurons\\per Layer\end{tabular}} &
\begin{tabular}{@{}c@{}}Out-degree\\per Node\end{tabular} &
\begin{tabular}{@{}c@{}}Accuracy\\(\%)\end{tabular} &
\begin{tabular}{@{}c@{}}Number of\\Parameters\end{tabular} \\
\midrule
Deep    & [800 & \bfseries 6 6   & 10] & [6 6 10]   & 90.68 &  4918 \\
Shallow & [800 & \bfseries 6     & 10] & [6 10]     & 91.03 &  4876 \\
Sparse  & [800 & \bfseries 40 40 & 10] & [5 11 10]  & 95.65 &  4930 \\\hline
Deep    & [800 & \bfseries 10 10 & 10] & [10 10 10] & 94.59 &  8230 \\
Shallow & [800 & \bfseries 10    & 10] & [10 10]    & 94.00 &  8120 \\
Sparse  & [800 & \bfseries 40 40 & 10] & [9 14 10]  & 96.24 &  8250 \\\hline
Deep    & [800 & \bfseries 14 14 & 10] & [14 14 10] & 95.43 & 11574 \\
Shallow & [800 & \bfseries 14    & 10] & [14 10]    & 94.88 & 11364 \\
Sparse  & [800 & \bfseries 40 40 & 10] & [13 17 10] & 96.72 & 11570 \\\hline
Deep    & [800 & \bfseries 28 28 & 10] & [28 28 10] & 96.79 & 23530 \\
Shallow & [800 & \bfseries 29    & 10] & [29 10]    & 96.79 & 23529 \\
Sparse  & [800 & \bfseries 80 80 & 10] & [26 22 10] & 97.33 & 23530 \\\hline
Deep    & [800 & \bfseries 39 39 & 10] & [39 39 10] & 97.35 & 33199 \\
Shallow & [800 & \bfseries 41    & 10] & [41 10]    & 97.34 & 33261 \\
Sparse  & [800 & \bfseries 80 80 & 10] & [37 33 10] & 97.75 & 33210 \\
\bottomrule
\end{tabular}%

%% file: graphs/accuracy_head.tex
\begin{tikzpicture}
\begin{axis}[
    xlabel={Number of parameters},
    ylabel={Accuracy (\%)},
    xmin=2000, xmax=18000,
    ymin=75, ymax=100,
    xtick distance={3000},
    legend pos=south east,
    xmajorgrids=true,
    ymajorgrids=true,
    grid style=dashed,
    /pgf/number format/.cd,
    set thousands separator={},
    scaled ticks=false,
    tick label style={font=\small},
    xlabel style={yshift=+0.1cm},
    ylabel style={yshift=-0.2cm},
]
\addplot[color=blue, mark=square*] coordinates {
    ( 2450, 93.93)
    ( 4090, 95.30)
    ( 5730, 95.89)
    (11570, 96.54)
    (16650, 96.98)
};
\addlegendentry{Sparse}
\addplot[color=red, mark=*] coordinates {
    ( 2455, 80.89)
    ( 4065, 89.84)
    ( 5687, 92.22)
    (11364, 95.44)
    (17041, 96.03)
};
\addlegendentry{Shallow}
\addplot[color=green, mark=triangle*] coordinates {
    ( 2443, 76.58)
    ( 4095, 90.69)
    ( 5743, 92.85)
    (11574, 95.35)
    (16650, 96.03)
};
\addlegendentry{Deep}
\end{axis}
\end{tikzpicture}

%% file: graphs/accuracy_tail.tex
\begin{tikzpicture}
\begin{axis}[
    xlabel={Number of parameters},
    ylabel={Accuracy (\%)},
    xmin=4000, xmax=35000,
    ymin=75, ymax=100,
    xtick distance={5000},
    legend pos=south east,
    xmajorgrids=true,
    ymajorgrids=true,
    grid style=dashed,
    /pgf/number format/.cd,
    set thousands separator={},
    scaled ticks=false,
    tick label style={font=\small},
    xlabel style={yshift=+0.1cm},
    ylabel style={yshift=-0.2cm},
]
\addplot[color=blue, mark=square*] coordinates {
    ( 4930, 95.65)
    ( 8250, 96.24)
    (11570, 96.72)
    (23530, 97.33)
    (33210, 97.75)
};
\addlegendentry{Sparse}
\addplot[color=red, mark=*] coordinates {
    ( 4918, 90.68)
    ( 8230, 94.59)
    (11574, 95.43)
    (23530, 96.79)
    (33199, 97.35)
};
\addlegendentry{Shallow}
\addplot[color=green, mark=triangle*] coordinates {
    ( 4876, 91.03)
    ( 8120, 94.00)
    (11364, 94.88)
    (23529, 96.79)
    (33261, 97.34)
};
\addlegendentry{Deep}
\end{axis}
\end{tikzpicture}

%% file: srcs/5_conclusion.tex
\section{Conclusion}
\label{cha:conclusions}

In this paper, we presented the effect of predefined sparsity within the \gls{sc} and \gls{ee} paradigm.
This approach, demonstrably effective for the first time in an \gls{sc} and \gls{ee} scenario, significantly reduces the computational, storage, and energy demands during training and inference, regardless of the hardware platform.
The experimental results showcase impressive reductions exceeding $4\times{}$ in both storage and computational complexity while maintaining comparable accuracy.
This paves the way for deploying complex \gls{dnn} models on edge devices for real-world \gls{sc} and \gls{ee} applications.

%% file: 0_main.bbl
\begin{thebibliography}{10}
\providecommand{\url}[1]{#1}
\csname url@samestyle\endcsname
\providecommand{\newblock}{\relax}
\providecommand{\bibinfo}[2]{#2}
\providecommand{\BIBentrySTDinterwordspacing}{\spaceskip=0pt\relax}
\providecommand{\BIBentryALTinterwordstretchfactor}{4}
\providecommand{\BIBentryALTinterwordspacing}{\spaceskip=\fontdimen2\font plus
\BIBentryALTinterwordstretchfactor\fontdimen3\font minus \fontdimen4\font\relax}
\providecommand{\BIBforeignlanguage}[2]{{%
\expandafter\ifx\csname l@#1\endcsname\relax
\typeout{** WARNING: IEEEtran.bst: No hyphenation pattern has been}%
\typeout{** loaded for the language `#1'. Using the pattern for}%
\typeout{** the default language instead.}%
\else
\language=\csname l@#1\endcsname
\fi
#2}}
\providecommand{\BIBdecl}{\relax}
\BIBdecl

\bibitem{pouyanfar2018survey}
S.~Pouyanfar, S.~Sadiq, Y.~Yan \emph{et~al.}, ``A survey on deep learning: Algorithms, techniques, and applications,'' \emph{ACM Computing Surveys (CSUR)}, 2018.

\bibitem{dong2021survey}
S.~Dong, P.~Wang, and K.~Abbas, ``A survey on deep learning and its applications,'' \emph{Computer Science Review}, 2021.

\bibitem{smith2022using}
S.~Smith, M.~Patwary, B.~Norick \emph{et~al.}, ``Using deepspeed and megatron to train megatron-turing nlg 530b, a large-scale generative language model,'' \emph{arXiv preprint arXiv:2201.11990}, 2022.

\bibitem{chan2023deep}
K.~Y. Chan, B.~Abu-Salih, R.~Qaddoura \emph{et~al.}, ``Deep neural networks in the cloud: Review, applications, challenges and research directions,'' \emph{Neurocomputing}, 2023.

\bibitem{jouppi2017datacenter}
N.~P. Jouppi, C.~Young, N.~Patil \emph{et~al.}, ``In-datacenter performance analysis of a tensor processing unit,'' in \emph{Proceedings of the 44th annual international symposium on computer architecture}, 2017.

\bibitem{singh2023edge}
R.~Singh and S.~S. Gill, ``Edge ai: a survey,'' \emph{Internet of Things and Cyber-Physical Systems}, 2023.

\bibitem{capogrosso2024machine}
L.~Capogrosso, F.~Cunico, D.~S. Cheng \emph{et~al.}, ``A machine learning-oriented survey on tiny machine learning,'' \emph{IEEE Access}, 2024.

\bibitem{menghani2023efficient}
G.~Menghani, ``Efficient deep learning: A survey on making deep learning models smaller, faster, and better,'' \emph{ACM Computing Surveys}, 2023.

\bibitem{fragkou2023model}
E.~Fragkou, M.~Koultouki, and D.~Katsaros, ``Model reduction of feed forward neural networks for resource-constrained devices,'' \emph{Applied Intelligence}, 2023.

\bibitem{eshratifar2019jointdnn}
A.~E. Eshratifar, M.~S. Abrishami, and M.~Pedram, ``Jointdnn: An efficient training and inference engine for intelligent mobile cloud computing services,'' \emph{IEEE Transactions on Mobile Computing}, 2019.

\bibitem{jankowski2020joint}
M.~Jankowski, D.~G{\"u}nd{\"u}z, and K.~Mikolajczyk, ``Joint device-edge inference over wireless links with pruning,'' in \emph{2020 IEEE 21st International Workshop on Signal Processing Advances in Wireless Communications (SPAWC)}.\hskip 1em plus 0.5em minus 0.4em\relax IEEE, 2020.

\bibitem{sbai2021cut}
M.~Sbai, M.~R.~U. Saputra, N.~Trigoni \emph{et~al.}, ``Cut, distil and encode (cde): Split cloud-edge deep inference,'' in \emph{2021 18th Annual IEEE International Conference on Sensing, Communication, and Networking (SECON)}.\hskip 1em plus 0.5em minus 0.4em\relax IEEE, 2021.

\bibitem{cunico2022split}
F.~Cunico, L.~Capogrosso, F.~Setti \emph{et~al.}, ``I-split: Deep network interpretability for split computing,'' in \emph{2022 26th International Conference on Pattern Recognition (ICPR)}.\hskip 1em plus 0.5em minus 0.4em\relax IEEE, 2022.

\bibitem{capogrosso2023split}
L.~Capogrosso, F.~Cunico, M.~Lora \emph{et~al.}, ``Split-et-impera: A framework for the design of distributed deep learning applications,'' in \emph{2023 26th International Symposium on Design and Diagnostics of Electronic Circuits and Systems (DDECS)}.\hskip 1em plus 0.5em minus 0.4em\relax IEEE, 2023.

\bibitem{matsubara2022split}
Y.~Matsubara, M.~Levorato, and F.~Restuccia, ``Split computing and early exiting for deep learning applications: Survey and research challenges,'' \emph{ACM Computing Surveys}, 2022.

\bibitem{gong2014compressing}
Y.~Gong, L.~Liu, M.~Yang \emph{et~al.}, ``Compressing deep convolutional networks using vector quantization,'' \emph{arXiv preprint arXiv:1412.6115}, 2014.

\bibitem{chen2015compressing}
W.~Chen, J.~Wilson, S.~Tyree \emph{et~al.}, ``Compressing neural networks with the hashing trick,'' in \emph{International conference on machine learning}.\hskip 1em plus 0.5em minus 0.4em\relax PMLR, 2015.

\bibitem{wen2016learning}
W.~Wen, C.~Wu, Y.~Wang \emph{et~al.}, ``Learning structured sparsity in deep neural networks,'' \emph{Advances in neural information processing systems}, 2016.

\bibitem{srivastava2014dropout}
N.~Srivastava, G.~Hinton, A.~Krizhevsky \emph{et~al.}, ``Dropout: a simple way to prevent neural networks from overfitting,'' \emph{The journal of machine learning research}, 2014.

\bibitem{labach2019survey}
A.~Labach, H.~Salehinejad, and S.~Valaee, ``Survey of dropout methods for deep neural networks,'' \emph{arXiv preprint arXiv:1904.13310}, 2019.

\bibitem{albericio2016cnvlutin}
J.~Albericio, P.~Judd, T.~Hetherington \emph{et~al.}, ``Cnvlutin: Ineffectual-neuron-free deep neural network computing,'' \emph{ACM SIGARCH Computer Architecture News}, 2016.

\bibitem{aghasi2017net}
A.~Aghasi, A.~Abdi, N.~Nguyen \emph{et~al.}, ``Net-trim: Convex pruning of deep neural networks with performance guarantee,'' \emph{Advances in neural information processing systems}, 2017.

\bibitem{han2015deep}
S.~Han, H.~Mao, and W.~J. Dally, ``Deep compression: Compressing deep neural networks with pruning, trained quantization and huffman coding,'' \emph{arXiv preprint arXiv:1510.00149}, 2015.

\bibitem{sindhwani2015structured}
V.~Sindhwani, T.~Sainath, and S.~Kumar, ``Structured transforms for small-footprint deep learning,'' \emph{Advances in Neural Information Processing Systems}, 2015.

\bibitem{dey2019pre}
S.~Dey, K.-W. Huang, P.~A. Beerel \emph{et~al.}, ``Pre-defined sparse neural networks with hardware acceleration,'' \emph{IEEE Journal on Emerging and Selected Topics in Circuits and Systems}, 2019.

\bibitem{howard2019searching}
A.~Howard, M.~Sandler, G.~Chu \emph{et~al.}, ``Searching for mobilenetv3,'' in \emph{Proceedings of the IEEE/CVF international conference on computer vision}, 2019.

\bibitem{liang2021pruning}
T.~Liang, J.~Glossner, L.~Wang \emph{et~al.}, ``Pruning and quantization for deep neural network acceleration: A survey,'' \emph{Neurocomputing}, 2021.

\bibitem{gou2021knowledge}
J.~Gou, B.~Yu, S.~J. Maybank \emph{et~al.}, ``Knowledge distillation: A survey,'' \emph{International Journal of Computer Vision}, 2021.

\bibitem{matsubara2019distilled}
Y.~Matsubara, S.~Baidya, D.~Callegaro \emph{et~al.}, ``Distilled split deep neural networks for edge-assisted real-time systems,'' in \emph{Proceedings of the 2019 Workshop on Hot Topics in Video Analytics and Intelligent Edges}, 2019.

\bibitem{kang2017neurosurgeon}
Y.~Kang, J.~Hauswald, C.~Gao \emph{et~al.}, ``Neurosurgeon: Collaborative intelligence between the cloud and mobile edge,'' \emph{ACM SIGARCH Computer Architecture News}, 2017.

\bibitem{li2018auto}
G.~Li, L.~Liu, X.~Wang \emph{et~al.}, ``Auto-tuning neural network quantization framework for collaborative inference between the cloud and edge,'' in \emph{Artificial Neural Networks and Machine Learning--ICANN 2018: 27th International Conference on Artificial Neural Networks, Rhodes, Greece, October 4-7, 2018, Proceedings, Part I 27}.\hskip 1em plus 0.5em minus 0.4em\relax Springer, 2018.

\bibitem{choi2018deep}
H.~Choi and I.~V. Baji{\'c}, ``Deep feature compression for collaborative object detection,'' in \emph{2018 25th IEEE International Conference on Image Processing (ICIP)}.\hskip 1em plus 0.5em minus 0.4em\relax IEEE, 2018.

\bibitem{carra2023dnn}
D.~Carra and G.~Neglia, ``Dnn split computing: Quantization and run-length coding are enough,'' in \emph{GLOBECOM 2023-2023 IEEE Global Communications Conference}.\hskip 1em plus 0.5em minus 0.4em\relax IEEE, 2023.

\bibitem{eshratifar2019bottlenet}
A.~E. Eshratifar, A.~Esmaili, and M.~Pedram, ``Bottlenet: A deep learning architecture for intelligent mobile cloud computing services,'' in \emph{2019 IEEE/ACM International Symposium on Low Power Electronics and Design (ISLPED)}.\hskip 1em plus 0.5em minus 0.4em\relax IEEE, 2019.

\bibitem{caruana1997multitask}
R.~Caruana, ``Multitask learning,'' \emph{Machine learning}, 1997.

\bibitem{capogrosso2024mtl}
L.~Capogrosso, E.~Fraccaroli, S.~Chakraborty \emph{et~al.}, ``Mtl-split: Multi-task learning for edge devices using split computing,'' \emph{arXiv preprint arXiv:2407.05982}, 2024.

\bibitem{lo2017dynamic}
C.~Lo, Y.-Y. Su, C.-Y. Lee \emph{et~al.}, ``A dynamic deep neural network design for efficient workload allocation in edge computing,'' in \emph{2017 IEEE International Conference on Computer Design (ICCD)}.\hskip 1em plus 0.5em minus 0.4em\relax IEEE, 2017.

\bibitem{turetta2023towards}
C.~Turetta, G.~Skenderi, L.~Capogrosso \emph{et~al.}, ``Towards deep learning-based occupancy detection via wifi sensing in unconstrained environments,'' in \emph{2023 Design, Automation \& Test in Europe Conference \& Exhibition (DATE)}.\hskip 1em plus 0.5em minus 0.4em\relax IEEE, 2023.

\bibitem{chen2023tsmixer}
S.-A. Chen, C.-L. Li, N.~Yoder \emph{et~al.}, ``Tsmixer: An all-mlp architecture for time series forecasting,'' \emph{arXiv preprint arXiv:2303.06053}, 2023.

\bibitem{deng2012mnist}
L.~Deng, ``The mnist database of handwritten digit images for machine learning research,'' \emph{IEEE Signal Processing Magazine}, 2012.

\bibitem{deng2009imagenet}
J.~Deng, W.~Dong, R.~Socher \emph{et~al.}, ``Imagenet: A large-scale hierarchical image database,'' in \emph{2009 IEEE conference on computer vision and pattern recognition}.\hskip 1em plus 0.5em minus 0.4em\relax IEEE, 2009.

\bibitem{tensorflow2015}
\BIBentryALTinterwordspacing
M.~Abadi, A.~Agarwal, P.~Barham \emph{et~al.}, ``{TensorFlow}: Large-scale machine learning on heterogeneous systems,'' 2015, software available from tensorflow.org. [Online]. Available: \url{https://www.tensorflow.org/}
\BIBentrySTDinterwordspacing

\bibitem{kingma2014adam}
D.~P. Kingma and J.~Ba, ``Adam: A method for stochastic optimization,'' \emph{arXiv preprint arXiv:1412.6980}, 2014.

\end{thebibliography}
